\documentclass{article}

\usepackage{arxiv}

\usepackage[utf8]{inputenc} % allow utf-8 input
\usepackage[T1]{fontenc}    % use 8-bit T1 fonts
\usepackage{hyperref}       % hyperlinks
\usepackage{url}            % simple URL typesetting
\usepackage{booktabs}       % professional-quality tables
\usepackage{amsfonts}       % blackboard math symbols
\usepackage{nicefrac}       % compact symbols for 1/2, etc.
\usepackage{microtype}      % microtypography
\usepackage{lipsum}
\usepackage{graphicx}
\graphicspath{ {./images/} }
\usepackage{amsmath}
 
\title{A Lightweight Face Quality Assessment Framework
to Improve Face Verification Performance in Real-Time Screening Applications}

\author{
  Ahmed Aman Ibrahim\\
  theCircle Ltd \\
  London \\
  United Kingdom\\
  \texttt{ahmedamanibrahim@gmail.com} \\
  \And
  Hamad Mansour Alawar\\
  General Department of Forensic Science and Criminology \\
  Dubai Police Headquarters \\
  Dubai, United Arab Emirates\\
  \texttt{hm.alawar@dubaipolice.gov.ae} \\
  \And
  AbdulNasser Abbas Zehi\\
  General Department of Forensic Science and Criminology \\
  Dubai Police Headquarters \\
  Dubai, United Arab Emirates\\
  \texttt{aa.zehi@dubaipolice.gov.ae} \\
  \And
  Ahmed Mohammad Alkendi\\
  General Department of Forensic Science and Criminology \\
  Dubai Police Headquarters \\
  Dubai, United Arab Emirates\\
  \texttt{am.alkendi@dubaipolice.gov.ae} \\
   \And
  Bilal Shafi Ashfaq Ahmed Mirza\\
  theCircle Ltd \\
  London \\
  United Kingdom\\
  \texttt{bilalshafimirza@gmail.com} \\
   \And
  Shan Ullah\\
  theCircle Ltd \\
  London \\
  United Kingdom\\
  \texttt{shan.ullah.basir@gmail.com} \\
  \And
  Ismail Lujain Jaleel\\
  theCircle Ltd \\
  London \\
  United Kingdom\\
  \texttt{lujainjaleel@icloud.com} \\
   \And
  Hassan Ugail\\
  Centre for Visual Computing and Intelligent Systems \\
  Unviversity of Bradford \\
  United Kingdom\\
  \texttt{h.ugail@bradford.ac.uk} \\
}

\begin{document}
\maketitle

\begin{abstract}
Face image quality plays a critical role in determining the accuracy and reliability of face verification systems, particularly in real-time screening applications such as surveillance, identity verification, and access control. Low-quality face images, often caused by factors such as motion blur, poor lighting conditions, occlusions, and extreme pose variations, significantly degrade the performance of face recognition models, leading to higher false rejection and false acceptance rates. In this work, we propose a lightweight yet effective framework for automatic face quality assessment, which aims to pre-filter low-quality face images before they are passed to the verification pipeline. Our approach uses normalized landmarks in conjunction with a Random Forest Regression classifier to assess image quality, achieving an accuracy of 96.67\%. By integrating this quality assessment module into the face verification process, we observe a substantial improvement in performance, including a comfortable 99.7\% reduction in the false rejection rate and enhanced cosine similarity scores when paired with the ArcFace face verification model. To validate our approach, we have conducted experiments on a real-world dataset collected comprising over 600 subjects captured from CCTV footage in unconstrained environments within Dubai Police. Our results demonstrate that the proposed framework effectively mitigates the impact of poor-quality face images, outperforming existing face quality assessment techniques while maintaining computational efficiency. Moreover, the framework specifically addresses two critical challenges in real-time screening: variations in face resolution and pose deviations, both of which are prevalent in practical surveillance scenarios.
\end{abstract}

% keywords can be removed

\section{Introduction}\label{sec1}
Face recognition systems \cite{elmahmudi2019deep, goel2024sibling} have rapidly evolved into essential tools within the domain of surveillance, security, and identity verification. Their widespread deployment spans applications from border control and secure access systems to large-scale public surveillance networks. The appeal lies in the non-intrusive nature of face recognition, allowing identification of an individual at a distance without requiring active subject cooperation. However, despite the remarkable progress made by modern machine learning-based techniques, the reliability of face recognition systems in real-world, unconstrained environments remains a critical concern. A primary determinant of system performance is the quality of the input facial images.

While controlled settings, such as passport photo booths or security checkpoint cameras, can guarantee optimal image capture conditions, real time screening, such as surveillance scenarios, often presents a vastly different reality. CCTV systems, body-worn cameras, and mobile surveillance setups frequently operate under non-ideal conditions, characterized by varying pose angles, inconsistent illumination, low image resolution, motion blur, and partial occlusions. These factors collectively degrade input facial image quality, leading to diminished recognition accuracy and increasing the likelihood of false rejections and misidentifications. The implications are particularly concerning in high-stakes security operations where accurate as well as timely identification is of importance.

Modern face recognition architectures, such as ArcFace \cite{deng2019arcface} and FaceNet \cite{schroff2015facenet}, rely on deep embeddings derived from convolutional neural networks trained using triplet loss or angular margin loss to map facial images to discriminative feature representations. Recognition decisions are then based on distance metrics, such as cosine similarity, on these embeddings. While these models achieve state-of-the-art performance, surpassing 99\% accuracy on benchmark datasets like VGGFace2 \cite{cao2018vggface2} and LFW \cite{huang2008labeled}, their robustness often deteriorates when processing poor-quality images. Hence, embedding degradation results in increased intra-class variability and reduced inter-class separability, eroding recognition performance.

To address the quality issue in face recognition, researchers have proposed face quality assessment (FQA) frameworks \cite{TFace_SDD-FIQA2021, meng2021magface, faceqnet_hernandez2020biometric, terhorst2020ser, shi2019probabilistic, chang2020data}. These methods function as a preprocessing step, filtering out low-quality facial images before recognition is attempted. The core question underlying FQA research is: what defines a high-quality face image? The International Organization for Standardization (ISO) provides guidance through ISO/IEC TR 29794-5:2010 \cite{ISO29794-5:2010}, outlining key quality attributes such as pose, illumination, resolution, sharpness, facial expression, and occlusions. Complementary guidelines are provided by the Face Recognition Vendor Test (FRVT) from the National Institute of Standards and Technology (NIST) \cite{nist_frvt} and the International Civil Aviation Organization (ICAO) standards for passport photographs \cite{icao_doc9303}.

While these standards have been instrumental in defining face image quality for controlled environments, their applicability diminishes in surveillance contexts. Real-time screening scenarios often involve off-angle captures, faces partially obscured by masks or objects, and compressed, low-resolution video streams. These operational realities necessitate adaptive FQA frameworks capable of evaluating image quality under such unconstrained conditions, ensuring that face verification systems can function reliably in practical deployments.

\begin{figure}[!t]
\centering
\includegraphics[width=0.9\linewidth]{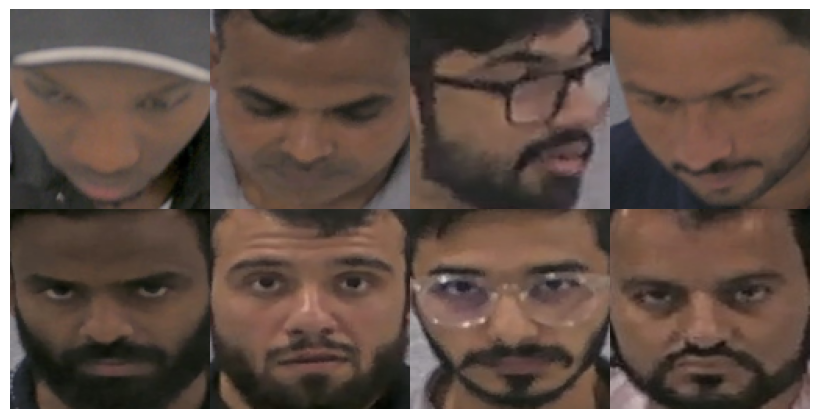}
\caption{Sample images from our dataset. Participants walked in the field of view of a CCTV camera. We then selected frames of low (top) and high (bottom) quality face images. High quality faces are determined using the face pose (face is towards the camera) and resolution (distance of participant is close to the camera).}
\label{fig:dataset_samples}
\end{figure}

To tackle these challenges, we collected a real-world surveillance dataset comprising CCTV footage of 600 participants \ref{fig:dataset_samples}. We manually labelled video frames containing faces as either high or low quality and developed a robust face quality assessment framework specifically engineered for surveillance environments. Our approach implements a three-stage pipeline combining face detection, facial landmark detection, and a machine learning classifier to evaluate face image quality before processing by face verification systems. The framework particularly addresses face resolution and pose variations, which are common challenges in CCTV environments.

The key contributions of this work include:
\begin{enumerate}
\item A comprehensive face quality assessment framework that leverages normalized facial landmark coordinates to evaluate image quality
\item An extensive comparative analysis of multiple classification approaches, including Logistic Regression, KNN, SVM, Random Forests, and Neural Networks, for face quality assessment and comparison with state-of-the-art face quality assessment frameworks
\item Quantitative evaluation demonstrating the framework's impact on face verification performance through cosine similarity analysis
\end{enumerate}

Our experimental results, based on a carefully curated dataset of approximately 600 subjects in real surveillance scenarios, demonstrate that incorporating our quality assessment framework as a preprocessing step significantly enhances face verification reliability. The proposed Random Forest-based model achieves superior performance to state-of-the-art methods on our dataset while being specifically optimized for surveillance applications.

The remainder of this paper is organised as follows. In Section \ref{method_}, we
 discuss our methodology, including dataset creation, model architecture, and normalisation techniques. In Section \ref{results_}, we present our experimental results and comparative analysis with existing approaches. In Section \ref{discussion_}, we discuss our findings and. Finally, in \ref{conclusion_}, we conclude the paper and explore potential future research directions of this work.

\section{Related Work}
Modern face recognition systems such as ArcFace \cite{deng2019arcface} and FaceNet \cite{schroff2015facenet} have achieved remarkable accuracy on benchmark datasets like LFW \cite{huang2008labeled} and YTF \cite{wolf2011face_ytf} through deep learning architectures and sophisticated loss functions. These systems transform facial images into high-dimensional embeddings that enable face comparison through distance metrics. While these models are trained to work on challenging environments like varying poses, occlusions and resolutions, the quality of face embeddings is still very much dependent on input face image quality \cite{shi2019probabilistic}. As \cite{faceqnet_hernandez2020biometric} aptly points out, this follows the principle of``garbage in, garbage out"—high-quality input face images produce high-quality embeddings, which in turn lead to better face recognition performance.

This raises the fundamental question: what defines a high-quality facial image? A simple definition given by \cite{grother2007performance} defines it as the utility of the image to the facial recognition system. If a face image is more suitable for a face recognition system, it is naturally of higher quality. However, we still need a way to measure high quality face images. While traditional image quality metrics \cite{eskicioglu1995image,sheikh2006image} measuring things such as resolution, illumination, blur, and noise levels—provide a starting point, face image quality assessment must also consider face-specific attributes like pose and occlusions. 

Currently, there is no universal consensus on what constitutes optimal face image quality \cite{faceqnet_hernandez2020biometric}. Some researchers have approached this challenge by collecting human annotations \cite{best2018learning}, as demonstrated in studies using the LFW database \cite{huang2008labeled} where quality ratings were derived from pairwise image comparisons. Industry standards also provide various frameworks for quality assessment. For instance, NIST's FRVT \cite{grother2014face} program includes specific evaluations of algorithms' ability to predict face recognition performance through quality scores. Similarly, the ICAO requirements for passport photographs \cite{doc20159303} establish strict criteria for face pose, expression, illumination uniformity, resolution, and absence of occlusions—all designed to ensure reliable processing of faces by automated border control systems. However, there is a lack of face quality datasets for live surveillance environments. That is why we collected and labelled our own dataset for this use case.

Machine learning models for face quality assessment generally fall into two categories: supervised and unsupervised learning methods. Supervised approaches rely on human annotators who provide quality labels for face images, which are then used to train models to predict image quality. Unsupervised methods, in contrast, don't require explicit quality scores. Instead, these models learn to evaluate image quality by identifying and analysing inherent patterns and characteristics within face images.

Notable works which use supervised learning include \cite{best2018learning} and \cite{zhang2017illumination}. In \cite{best2018learning}, human annotations of face quality are collected using face images from the LWF database. Their approach involved showing human annotators a pair of face images and asking them to select which image was of higher quality. Then, using the matrix completion approach proposed by \cite{yi2013inferring}, quality ratings for individual images are inferred. Finally, a support vector regression (SVR) model is trained to predict these quality ratings. In \cite{zhang2017illumination}, face images are collected and labelled for their illumination quality. Then, a convolutional neural network (CNN) model is trained to predict the illumination quality values.

Unsupervised methods like \cite{meng2021magface} do not use explicit quality labels. They rely on the fact that face recognition datasets such as LFW \cite{huang2008labeled} and MS-Celeb-1M \cite{guo2016ms} come with identity labels (who is in each image). A face verification model like ArcFace \cite{deng2019arcface} is then used to compare images of each person using a similarity metric like cosine similarity. The face verification model learns that some images of the same person are more easy to recognise than others. The images that are easy to recognize are then labelled as having higher quality.

Looking at how unsupervised learning methods generate quality labels, in \cite{meng2021magface}, an embedding method is introduced where the magnitude of the embedding monotonically increases for higher quality face images. In \cite{TFace_SDD-FIQA2021}, Wasserstein distance between inter-class (same subject) and inter-class (different subjects) face images are calculated to produce pseudo-quality labels. \cite{faceqnet_hernandez2020biometric} uses a classifier to identify ICAO-compliant \cite{icao_doc9303} reference images, assume they are of high quality and compare the similarity of the rest of the images to the high quality ones. \cite{terhorst2020ser} measures the embedding variations from random subnetworks of a face verification model to estimate quality. Once the quality labels are generated, either a regression model (generates a quality score) or a classification model (high or low quality) is trained.

Unsupervised methods offer the significant advantage of eliminating the need for costly human data labelling. However, they face several critical limitations. A primary concern is their reliance on face verification models like ArcFace, which, despite achieving an impressive 91.25\% overall recognition accuracy, still struggles with low-quality images. This creates a circular dependency problem: the very images that need quality assessment are often the ones where these models perform poorly, potentially leading to unreliable quality labels. Furthermore, research by \cite{best2018learning} demonstrates that human quality assessments strongly correlate with face recognition accuracy, while studies in cognitive neuroscience have shown that the human visual system possesses remarkable face processing capabilities \cite{kanwisher2006fusiform, blanton2016comparison}. Given these considerations, our method employs human-labeled face quality data focused on pose and resolution, leveraging human perceptual expertise while maintaining objective assessment criteria.

\section{Methodology}\label{method_}

\subsection{Dataset Creation}
To frame face quality assessment as a supervised learning problem, we developed a labelled dataset of facial images categorized as either high or low quality. The data collection process involved setting up a CCTV camera and recruiting volunteers to walk through its field of view. We captured facial images at various points during their walk. Images where the subject's face was fully visible to the camera were labelled as high quality, while those with partial or obscured facial views were designated as low quality. This systematic approach allowed us to create a comprehensive training dataset that represents real-world surveillance scenarios. We split this dataset into 80\% train and 20\% test sets. Since the dataset is privately owned by Dubai Police, we are not able to publish it; however, we display some sample images (Figure \ref{fig:dataset_samples}).

\subsection{Model}
After selecting relevant frames from the CCTV footage for our training dataset, we implemented a three-stage facial analysis pipeline (Figure \ref{fig:fqa_pipeline}). First, we employed RetinaFace \cite{deng2020retinaface} for face detection, followed by 5-point facial key point detection \cite{deng2019arcface} to extract precise facial key points corresponding to both eyes, the tip of the nose and the corners of the lips. The model outputs the x and y coordinates for each of the 5 key points. To account for scale variations caused by different distances between subjects and the camera, we normalized the key point coordinates relative to the face detection bounding box using the following formula:

\begin{figure*}[ht]
    \centering
    \includegraphics[width=\textwidth]{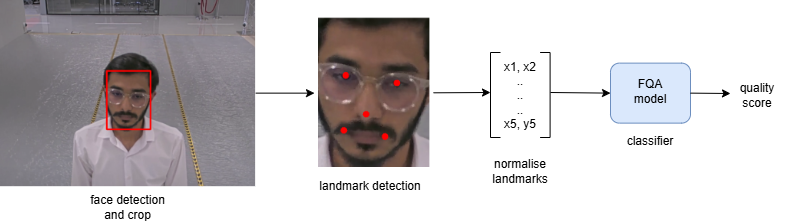}
    \caption{Our proposed face quality assessment pipeline. Face detection is performed first, and the face is cropped. Next, we detect the face key points and normalise them. The normalised face key points are then passed to the face quality assessment (FQA) model, which gives a quality score between 0 and 1.}
    \label{fig:fqa_pipeline}
\end{figure*}

\begin{equation}
x' = \frac{x - x_{\text{box}{\text{min}}}}{x_{\text{box}{\text{max}}} - x_{\text{box}{\text{min}}}}, \quad y' = \frac{y - y_{\text{box}{\text{min}}}}{y_{\text{box}{\text{max}}} - y_{\text{box}_{\text{min}}}},
\end{equation}

where $(x', y')$ represents the normalized coordinates, $(x, y)$ denotes the original key point coordinates, and $(x_{\text{box}{\text{min}}}, y_{\text{box}{\text{min}}})$ and $(x_{\text{box}{\text{max}}}, y_{\text{box}_{\text{max}}})$ represent the minimum and maximum coordinates of the bounding box respectively. Finally, these normalised coordinates are passed to the classifier model, which determines whether a face is high quality or not.

\subsection{Performance Evaluation}
Our evaluation strategy comprised two complementary approaches to validate the effectiveness of our face quality assessment framework. First, we conducted a standalone evaluation using standard classification metrics: accuracy, precision, recall, and F1 score. Second, we quantitatively assessed the framework's impact on face verification performance.
For the face verification experiments, we employed the ArcFace model \cite{deng2019arcface} as our baseline verification system. The experimental protocol consisted of the following steps:

For each subject, we stored a high-quality reference image and computed its facial embedding using ArcFace.
We evaluated face verification performance under two conditions:
\begin{itemize}
\item Baseline: Processing all CCTV-captured images without quality filtering
\item Proposed: Processing only images that passed our quality assessment framework
\end{itemize}

Face verification decisions were based on the cosine similarity between embeddings, defined as:
\begin{equation}
similarity(e_1, e_2) = \frac{e_1 \cdot e_2}{|e_1| |e_2|},
\end{equation}
where $e_1$ and $e_2$ represent the embeddings of the high quality reference image and probe images, respectively. We established a similarity threshold of 0.5, meaning face pairs with similarity scores above this threshold were considered matches, corresponding to 75\% similarity between embeddings (Figure).

To quantify the impact of quality filtering, we computed the following metrics:
\begin{itemize}
\item Mean and standard deviation of cosine similarity scores
\item False Rejection Rate (FRR), defined as:
\begin{equation}
FRR = \frac{\text{No. of incorrectly rejected genuine matches}}{\text{Total no. of genuine match attempts}}.
\end{equation}
\end{itemize}

\section{Results}\label{results_}
\subsection{Face Quality Assessment Performance}
We evaluated $5$ different classifier models on our face quality assessment dataset. While Logistic Regression and Neural Networks had a relatively poor performance, KNN, SVC and Random Forests all had an accuracy $>90$, with Random Forests having the best performance with an accuracy of 96.67\%. The results are summarised in Table \ref{tab:face_quality_model_comparison}.

\begin{table}[!t]
    \centering
    \caption{Face Quality Assessment Results for Different Classifier Models}
    \label{tab:face_quality_model_comparison}
    \begin{tabular}{l|cccc}
        \hline
        \textbf{Model} & \textbf{Accuracy} & \textbf{Precision} & \textbf{Recall} & \textbf{F1} \\
        \hline
        Logistic Regression & 73.33 & 79.07 & 69.39 & 73.91 \\
        Neural Network & 80.00 & 81.63 & 82.51 & 81.11 \\
        SVC & 90.00 & 84.48 & 100 & 91.59 \\
        KNN & 93.33 & 90.57 & 97.96 & 94.12 \\
        Random Forests & 96.67 & 97.92 & 95.91 & 96.93 \\
        \hline
    \end{tabular}
\end{table}

Subsequently, we compared our best-performing model (Random Forest) against existing approaches from the literature. Most existing methods demonstrated suboptimal performance, with the exception of SER-FIQ \cite{terhorst2020ser}, which achieved comparable results to our model.

\begin{table}[!t]
    \centering
    \caption{Face Quality Assessment Results, Our Model vs Literature}
    \label{tab:face_quality_literature_comparison}
    \begin{tabular}{l|cccc}
        \hline
        \textbf{Model} & \textbf{Accuracy} & \textbf{Precision} & \textbf{Recall} & \textbf{F1} \\
        \hline
         MagFace \cite{meng2021magface} & 12.22 & 21.15 & 22.45 & 21.78 \\
         TFace \cite{TFace_SDD-FIQA2021} & 77.78 & 93.94 & 63.27 & 75.61 \\
         FaceQNet v2 \cite{faceqnet_hernandez2020biometric} & 80.00 & 96.97 & 65.31 & 78.05 \\
         SER-FIQ \cite{terhorst2020ser} & 91.95 & 88.46 & 97.87 & 92.93 \\
         Random Forests (Ours) & 96.67 & 97.92 & 95.92 & 96.91 \\
        \hline
    \end{tabular}
\end{table}

\subsection{Impact on Face Verification}
The integration of our quality assessment framework demonstrated significant improvements in face verification performance. When comparing the baseline approach (processing all images) against our proposed quality-filtered approach, we observed substantial enhancements across key metrics (Table \ref{tab:impact_face_verification}).

The mean cosine similarity between face pairs increased from $0.66$ in the baseline scenario to 0.76 with quality filtering, indicating stronger matches between genuine pairs. This 15\% improvement in similarity scores suggests that our framework effectively identifies and retains the highest quality facial images that are most suitable for verification.

Most notably, the False Rejection Rate (FRR) showed dramatic improvement, decreasing from 13.19\% when processing all images to just 0.04\% when employing our quality filtering approach. This represents a 99.7\% reduction in false rejections, meaning that genuine subjects are now rarely misclassified as impostors. 

\begin{table}[!t]
\caption{Model Performance Comparison}
\label{tab:impact_face_verification}
\centering
\begin{tabular}{|l|c|c|}
\hline
\textbf{} & \textbf{Mean Cosine Similarity} & \textbf{FRR (\%)} \\
\hline
All images & 0.66 & 13.19 \\
With quality filtering & 0.76 & 0.04 \\
\hline
\end{tabular}
\end{table}

\section{Discussion}\label{discussion_}
Our approach to face quality assessment involved computing and normalizing facial key points, which were then fed into a machine learning model along with human-labeled quality assessments. The comparison of different machine learning methods (Table \ref{tab:face_quality_model_comparison}) revealed that face quality assessment using facial key points exhibits strongly non-linear characteristics, as evidenced by the poor performance of logistic regression (accuracy $<73.33$\%). Non-linear methods, including SVC, KNN, and Random Forests, demonstrated superior performance ($>90$\% accuracy), with the exception of neural networks. The relatively poor performance of neural networks can likely be attributed to our dataset's modest size (600 subjects), which made the model prone to overfitting.

When compared to state-of-the-art approaches, our facial quality assessment framework achieved exceptional performance while substantially improving the reliability of face verification systems in surveillance scenarios. Our Random Forest classifier achieved 96.67\% accuracy, outperforming most existing approaches (Table \ref{tab:face_quality_literature_comparison}). This superior performance can be attributed to two key factors. First, our use of normalized 2D-106-point facial landmarks provides a more comprehensive representation of facial geometry compared to traditional approaches that rely solely on image-based features. Second, our dataset's focus on real-world CCTV scenarios better captures the specific challenges encountered in surveillance applications, whereas existing methods are typically optimized for more controlled environments. Notably, only SER-FIQ achieved comparable performance to our model, suggesting its robust applicability across diverse scenarios.

The integration of our quality assessment framework with ArcFace for face verification yielded remarkable improvements in performance. The False Rejection Rate (FRR) dropped dramatically from 13.19\% to 0.04\% (Table \ref{tab:impact_face_verification}), representing a 99.7\% reduction in false rejections. This improvement directly addresses one of the primary challenges in surveillance-based face verification systems: the risk of misidentifying authorized individuals due to poor-quality images. Additionally, the increase in mean cosine similarity from 0.66 to 0.76 demonstrates that face verification models produce more confident and higher quality embeddings when images are filtered for quality. This correlation between human-annotated quality assessments and improved face verification performance aligns with findings from \cite{best2018learning}.

Despite these promising results, our study has several limitations that warrant acknowledgement. First, while comprehensive for our specific use case, our dataset is privately owned and relatively small, comprising approximately 600 subjects. As demonstrated by the performance results in Table \ref{tab:face_quality_model_comparison}, this dataset size proves insufficient for training deep learning models effectively. Additionally, our focus on face resolution and pose, while important, may limit the generalisability of our findings to other surveillance contexts where factors such as occlusions play a significant role.

\section{Conclusion}\label{conclusion_}
This work presents a robust face quality assessment framework for surveillance environments. We first collected and annotated a dataset of 600 subjects, categorizing faces as high or low quality. The framework operates through face detection and key point detection, using the face detection bounding box area to determine resolution. High-resolution faces are filtered by passing normalized face key points into a classifier model trained on human-supervised quality labels. Our Random Forest approach achieved 96.67\% accuracy, surpassing existing deep learning methods. Integration with ArcFace significantly improved face verification performance, reducing the False Rejection Rate from 13.19\% to 0.04\% and lowering the risk of misidentifying known individuals. The mean cosine similarity increased from 0.66 to 0.76, indicating that quality-filtered faces produce better embeddings and enable more confident model predictions.

While our results are promising, limitations include the relatively small dataset size of 600 subjects and focus primarily on face resolution and pose variations. Future work should explore expanding the framework to handle additional quality factors such as occlusions and extreme lighting conditions. Additionally, investigating the application of deep learning approaches with larger datasets could potentially yield further improvements in quality assessment accuracy.

Our findings demonstrate that incorporating quality assessment as a preprocessing step significantly enhances face verification reliability in surveillance applications. This work contributes to the broader goal of improving real-world face recognition systems by establishing effective quality thresholds for unconstrained environments such as live CCTV surveillance.

\section*{Data Availability}\label{data_}
The datasets utilized for training and evaluation are available on request (in
anonymized format).

\section*{Ethics declarations}\label{declarations_}
{\bf Consent to Participate and Consent to Publish:} \\
Informed consent was obtained from all individual participants included in the study and permission has been sought to publish the relevant information and data. \\

\flushleft {\bf Conflict of interest:} \\
The authors have no competing interests to declare that are relevant to the content of this article.

\section*{Funding}\label{fundning_}
This work was supported by a Dubai Police R\&D project  - Project Number P4864.

\section*{Contributions}\label{contributions_}
HU led the research, while AAI and BSAAM conducted the experiments. AAI and BSAAM designed the algorithms, executed the experiments, and generated the results. All authors discussed the experimental findings. AAI drafted the manuscript, HU, SU, HMA, AMA and AAZ reviewed and finalized it.

\end{document}